\documentclass{article}
\usepackage{ijcai24}

\usepackage{times}
\usepackage{soul}
\usepackage{url}
\usepackage[hidelinks]{hyperref}
\usepackage[utf8]{inputenc}
\usepackage[small]{caption}
\usepackage{graphicx}
\usepackage{amsmath}
\usepackage{amsthm}
\usepackage{booktabs}
\usepackage{algorithm}
\usepackage{algorithmic}
\usepackage[switch]{lineno}

\usepackage{verbatim}
\usepackage{multirow}
\usepackage{subfigure}
\usepackage{amsmath}
\usepackage{amssymb}
\usepackage{mathrsfs}
\usepackage{graphicx}
\usepackage{adjustbox}

\author{
Yonghao Liu$^1$
\and
Mengyu Li$^{1}$\and
Di Liang$^{2}$\and
Ximing Li$^1$ \and
Fausto Giunchiglia$^3$ \and \\
Lan Huang$^{1}$ \and
Xiaoyue Feng$^{1\ast}$ \And
Renchu Guan$^{1}$\thanks{Corresponding Author}
\affiliations
$^1$Key Laboratory of Symbolic Computation and Knowledge Engineering of the Ministry of \\ Education, College of Computer Science and Technology, Jilin University \\
$^2$Fudan University \\
$^3$University of Trento
\emails
\{yonghao20, mengyul21\}@mails.jlu.edu.cn,\\
liximing86@gmail.com, fausto.giunchiglia@unitn.it \\
\{huanglan, fengxy, guanrenchu\}@jlu.edu.cn
}

\title{Resolving Word Vagueness with Scenario-guided \\Adapter for Natural Language Inference}

\begin{document}

\maketitle

\begin{abstract}
Natural Language Inference (NLI) is a crucial task in natural language processing that involves determining the relationship between two sentences, typically referred to as the premise and the hypothesis. However, traditional NLI models solely rely on the semantic information inherent in independent sentences and lack relevant situational visual information, which can hinder a complete understanding of the intended meaning of the sentences due to the ambiguity and vagueness of language. 
To address this challenge, we propose an innovative \textbf{ScenaFuse} adapter that simultaneously integrates large-scale pre-trained linguistic knowledge and relevant visual information for NLI tasks. Specifically, we first design an image-sentence interaction module to incorporate visuals into the attention mechanism of the pre-trained model, allowing the two modalities to interact comprehensively. Furthermore, we introduce an image-sentence fusion module that can adaptively integrate visual information from images and semantic information from sentences. 
By incorporating relevant visual information and leveraging linguistic knowledge, our approach bridges the gap between language and vision, leading to improved understanding and inference capabilities in NLI tasks. Extensive benchmark experiments demonstrate that our proposed ScenaFuse, a scenario-guided approach, consistently boosts NLI performance. 
\end{abstract}

\section{Introduction}
``\textit{Humans rely less on words than on visual images, auditory images and propositions or rules of logic in order to think. Even commonly used metaphors and analogies employ visual and spatial attributes to provide us with a quick and easy context in which to communicate and build thoughts \cite{pinker2003mind}},'' claimed a world-renowned linguist and evolutionary psychologist, \textit{Steven Pinker}. This statement highlights the potential significance of visual information in human cognitive processing of real-world challenges, suggesting that it may, in fact, outweigh the importance of symbolic natural language information. This proposition 
warrants further investigation and consideration within the context of cognitive science and human-computer interaction research.

Natural language inference (NLI), as a vital task in NLP, has received widespread attention since it is closely related to human language comprehension and reasoning abilities. NLI involves knowledge from multiple fields, such as natural language understanding \cite{storks2019recent,liu2024improved} and logical reasoning \cite{li2024simple,liu2024simple}, and is a challenging task. It has been widely used in various practical application scenarios, such as question answering \cite{liulocal,liu2023time}, text summarization \cite{falke2019ranking,liu2021vplag,liu2021deep}, and information retrieval \cite{murty2021dreca,liu2022few}. 
Generally, NLI consists of two sentences, referred to as the premise and the hypothesis. The premise is an assertive sentence that provides descriptive information, while the hypothesis is a sentence that requires inference based on the premise. The goal of NLI is to determine whether the hypothesis can be inferred from the premise (\textit{i.e.}, entailment), or contradicts the premise (\textit{i.e.}, contradiction), or has no relationship with it (\textit{i.e.}, neutral). 

\begin{figure}
    \centering
    \includegraphics[width=0.5\textwidth]{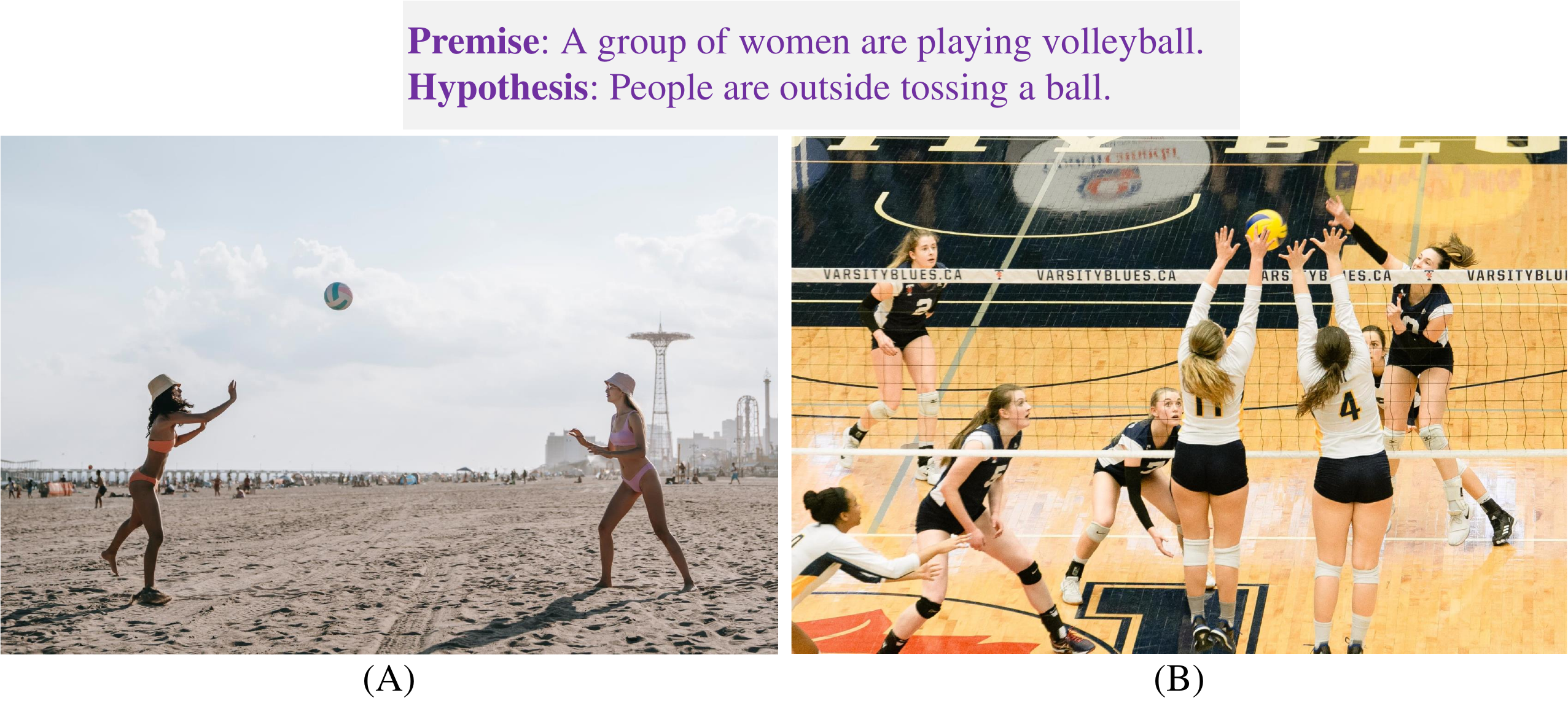}
    \caption{The two sentences have an entailment relationship when viewed in the context of image (A), but a contradiction relationship when viewed in the context of image (B).}
    \label{example}
\end{figure}
Recently, modern NLI models \cite{chen2017enhanced,conneau2017supervised} have made significant strides by leveraging the powerful feature extraction capabilities of deep learning to enhance their ability to understand and represent the semantic information of sentences. Successfully determining the semantic relationship in NLI tasks requires a deep comprehension of the semantics between sentences. 
Moreover, to further promote the development of NLI, researchers use human-labeling approaches to obtain large-scale high-quality annotated datasets, such as SNLI \cite{bowman2015large} and MultiNLI \cite{williams2018broad}, to evaluate the effectiveness of the proposed models. Despite the fruitful success achieved on these datasets, most previous models are still limited in performance due to their focus on extracting textual information while ignoring other useful information, such as visuals, which are crucial for proper sentence semantics understanding 
that can be used to discern the semantic relations between sentences. 

Previous studies \cite{gibson2019efficiency} have highlighted the limitations of language due to its inherent ambiguity and vagueness. This often leads to challenges in interpretation, as evident in the SNLI dataset, where different annotators may perceive the same pair of sentences as either an entailment or a contradiction. In contrast, 
images are more concrete and can convey complementary information \cite{pan2016jointly} that is often overlooked by sentences such as facial expressions and motions, which can effectively solve textual ambiguity \cite{tong2020image}. Thus, exploring the complementary information conveyed by images can help gain better insights into actual sentence meanings. 
As shown in Fig. \ref{example}, both the premise and hypothesis describe the scenario of people playing with balls. However, while the hypothesis specifies an outdoor setting, the location in the premise remains ambiguous. This ambiguity indicates a neutral relationship between the premise and hypothesis. 
Once the image related to the premise is provided, the relationship between the premise and hypothesis can be determined with certainty. In Fig. \ref{example} (A), the location information suggests that people are outdoors, creating an entailment relationship between the premise and hypothesis. Conversely, in Fig. \ref{example} (B), the location information indicates people are indoors, resulting in a contradictory relationship between the two. From this example, we can conclude that scenario images further enrich the semantics of sentences and provide necessary supplementary information. 
Although many images are gathered alongside the sentences during dataset collection, they are not given much attention. Hence, it is necessary to leverage relevant scenario images to fully understand the sentence semantics for NLI tasks.

Our objective is to incorporate scenario information into our sentence-level representations in a precise manner, which poses a considerable challenge. Naive integration of the two modalities can introduce noise that hinder semantic understanding, so finding an effective way to merge images with sentences remains an open problem. Although extensive research on multimodal tasks, there is currently no widely accepted method for combining image modality with sentences. 

To address the aforementioned challenges, we propose a novel scenario-guided adapter, coined \textbf{ScenaFuse}, for NLI tasks. Our approach involves designing an image-sentence interaction module that generates scenario-guided sentence semantic representations by deeply interacting images and sentences in the attention module of the pre-trained model. 
This module benefits the semantic understanding by explicitly considering the scenario images. Consequently, the text representations obtained from our approach incorporate valuable scenario information, are more comprehensive, and possess stronger representational power compared to those generated using a single modality. Moreover, we introduce an attention-based image-sentence fusion module with the gate mechanism to perform multimodal feature fusion related to sentence semantic and scenario-guided representations. This module selects informative features from both modalities while minimizing the noise impact, which facilitates the evaluation of sentence relationships. 
In summary, our contributions are listed as follows:

\noindent $\bullet$ We introduce a novel scenario-guided adapter named \textbf{ScenaFuse}, which explicitly incorporates visual information from the relevant scenarios. This is a valuable attempt at integrating scene information into large pre-trained language models for NLI tasks.

\noindent $\bullet$ We propose two modules: the image-sentence interaction module and the image-sentence fusion module. The former enables deep interaction between images and sentences, and the latter integrates two different modalities of features through specifically designed mechanisms. Moreover, we further demonstrate the effectiveness of the proposed approach on recent popular large language models.

\noindent $\bullet$ We conduct extensive experiments on benchmarks to evaluate the effectiveness of our approach. Our empirical results demonstrate the superiority of ScenaFuse compared to other competitive baselines. 

\section{Related Work}
\subsection{Natural Language Inference}
NLI models can be mainly divided into three categories: symbolic, statistical, and neural network-based models \cite{storks2019recent}. Symbolic models use logical forms to perform inference tasks and such models are heavily used in early NLI problems known as recognizing textual entailment (RTE) \cite{maccartney2009natural}. As data-driven feature training becomes more prevalent in NLP, statistical models based on feature engineering such as bag-of-words \cite{zhang2010understanding}, become the mainstream approach. Some work \cite{haim2006second,dagan2006pascal} also attempts to incorporate external knowledge as a supplement to training data features and achieve favorable results in NLI. The recent advancement in deep learning, along with the availability of larger annotated datasets, has led to the success of more complex neural network models in NLI tasks. These models can be categorized into two types: sentence-encoding-based and inter-sentence-attention-based. Sentence-encoding-based models \cite{conneau-etal-2017-supervised} utilize the same neural network architecture, such as LSTM \cite{bowman2015large} and GRU \cite{vendrov2015order} and their variants, to encode both the premise and hypothesis. A neural network classifier is then utilized to predict the relationship between the two sentences. On the other hand, models based on inter-sentence attention \cite{rocktaschel2015reasoning,wang2016learning} introduce attention mechanisms between the premise and hypothesis. This not only alleviates the vanishing gradient problem but also provides alignment between inputs and outputs, which allows for a better understanding of their relationships. 
\subsection{Multimodal Learning}
Multimodal data has garnered considerable attention from both academia and industry due to its potential applications in learning, analysis, and research \cite{ramachandram2017deep}. Multimodal models aim to extract information from varying modalities to learn expressive features that can be used for downstream tasks. The core of multimodal learning is to map unimodal data into a multimodal space, and subsequently acquiring the joint representation. Classic methods involve concatenating features from each modality and training the model, while more advanced approaches utilize neural networks to input each modality into several separate neural layers in an end-to-end manner, and then project them into hidden layers in the joint space \cite{antol2015vqa}. This joint multimodal representation can then be processed through multiple hidden layers or directly used for prediction. Due to the incorporation of large quantities of data during neural network training, the resulting joint representation typically performs well on various tasks. Recently, many studies have introduced multimodality into NLP tasks, such as named entity recognition \cite{moon2018multimodal} and machine translation \cite{nishihara2020supervised}. Generally, these methods enhance rough text understanding from non-textual perspectives by employing modality attention mechanisms to integrate information from heterogeneous sources. For example, MNMT \cite{nishihara2020supervised} incorporates a supervised visual attention mechanism, which is trained with constraints between manually aligned words in the sentence and corresponding regions in the image. This mechanism can more accurately capture the relationship between words and image regions. MNER \cite{moon2018multimodal}, on the other hand, proposes a universal modality-attention module that learns to reduce irrelevant modalities while amplifying the most informative ones for named entity recognition of data composed of text and images. 

\section{Problem Definition}
    
In this section, we define the NLI problem in detail. Consider a scenario where we have been provided with a premise consisting of $m$ words, represented by $P=(w_{p,1}, w_{p,2}, \cdots, w_{p,m})$, as well as a hypothesis with $n$ words, depicted by $H=(w_{h,1}, w_{h,2}, \cdots, w_{h,n})$. Additionally, an associated image $I$ is also available. 
The token embeddings $\mathbf{X}_{tex}$ in $P$ and $H$ are the sum of word, segment and position embeddings for each token, where the word embeddings can be initialized randomly or using word2vec or GloVe, \textit{i.e.}, $\mathbf{X}_{tex,i} \in \mathbb{R}^d$. In general, we follow the experimental settings of large-scale pre-trained models such as BERT \cite{devlin2019bert} or RoBERTa \cite{liu2019roberta}. To formalize two sentences into this format, we use the sentence pair ``[CLS] $P$ [SEP] $H$ [SEP]'' as model inputs. Here, [CLS] is a special symbol added to the beginning of each input example, and [SEP] is used as a special separator to separate sentences. We extract the [CLS] token as the output of each example. Because it lacks clear semantic information and can better integrate the semantic information of each word in the sentence. As a result, it represents the sentence's semantics better. Subsequently, we feed it into a fully connected layer for classification tasks. 

For the associated image $I$, we reshape it 224x224 pixels and input it into the pre-trained image encoder Enc($\cdot$), such as  VGG \cite{simonyan2014very} or ResNet \cite{he2016deep}, which have already demonstrated the ability to extract meaningful representations from input images in deep layers. Subsequently, we obtain its visual representations by retaining the output of the last convolutional layer. This process essentially segments each input image into visual blocks of the same size represented by $d^\prime$-dimensional vectors, denoted as $\mathbf{X}_{vis}=\text{Enc}(I)=(\mathbf{X}_{vis,1}, \mathbf{X}_{vis,2}, \cdots, \mathbf{X}_{vis,k})\in\mathbb{R}^{k\times d^\prime}$, where $\mathbf{X}_{vis,i}$ denotes the feature representation of the $i$-th visual block. 

In summary, we expect to train a model $\mathcal{F}$ that can precisely determine the relationship between the premise and the hypothesis when relevant data is provided, \textit{i.e.}, $y=\mathcal{F}(P, H, I) \in \{0, 1, 2\}$, where $y$=0 implies an entailment relationship, $y$=1 represents a neutral relationship, and $y$=2 represents a contradiction relationship.
\section{Method}
\begin{figure*}
    \centering
    \includegraphics[width=0.9\textwidth]{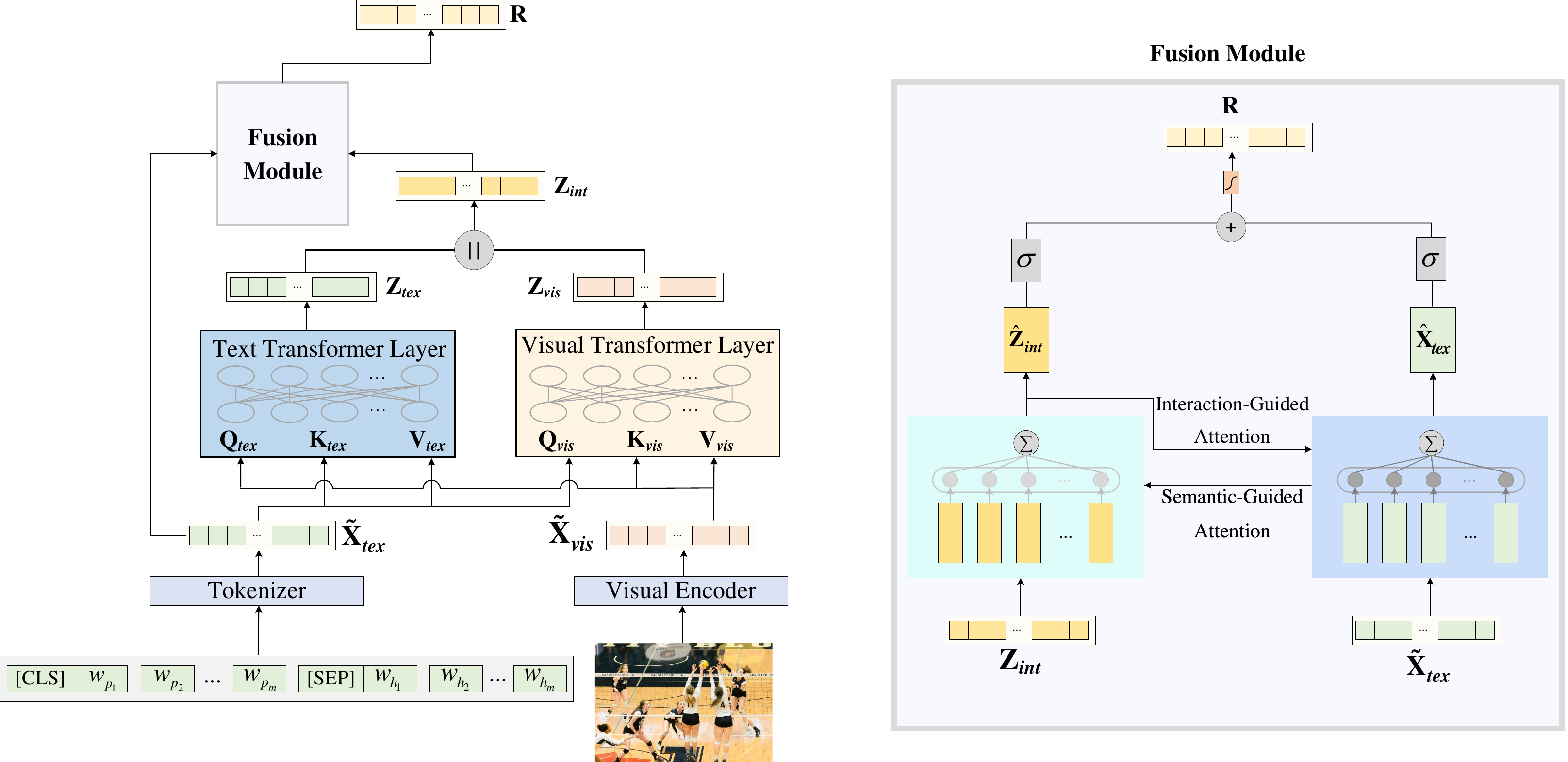}
    \caption{(Left) Overall architecture of our framework. (Right) The architecture of the fusion module. (Best viewed in color) }
    \label{framework}
\end{figure*}

This section provides an in-depth explanation of the two main modules that constitute the scenario-guided adapter, namely the image-sentence interaction and the image-sentence fusion. To facilitate a better understanding of our model, we have depicted its overall framework in Fig. \ref{framework}.
\subsection{Image-sentence Interaction}
As stated in the abstract, compared to language, images provide a more precise means of conveying information and may include important details that are difficult to express or neglected by language due to its inherent ambiguity and vagueness. Moreover, multimodal information has the potential to provide more discriminative input than a single modality, which can help address the issue of semantic ambiguity in text-only modality. Therefore, we explicitly incorporate visual contextual information to significantly enrich the semantic content of the sentence. To this end, we design an image-sentence interaction module. Specifically, we first adopt a linear transformation to project both modalities into the same representation space to achieve embedding space consistency and cross-modal semantic alignment between text representation modality $\mathbf{X}_{tex}$ and visual representation modality $\mathbf{X}_{vis}$. This can be expressed as follows: 
\begin{equation}
\begin{aligned}
    \tilde{\mathbf{X}}_{tex} = \varphi(\mathbf{X}_{tex}) \\
    \tilde{\mathbf{X}}_{vis} = \psi(\mathbf{X}_{vis})
\end{aligned}
\end{equation}
where $\varphi(\cdot)$ and $\psi(\cdot)$ are linear projection functions that project the text representation and visual representation to the same generated space, \textit{i.e.}, $\varphi: \mathbb{R}^{d}\rightarrow\mathbb{R}^{f}$ and $\psi: \mathbb{R}^{d^\prime}\rightarrow\mathbb{R}^{f}$. $\tilde{\mathbf{X}}_{tex}$ and $\tilde{\mathbf{X}}_{vis}$ are generated sentence representations and visual representations, respectively.

\paragraph{Visual-enhanced Sentence Representation.}
Subsequently, we consider that the textual content in NLI tasks is limited. To address this limitation and enable the utilization of visual information to enhance word representation learning, we employ a multi-head attention mechanism akin to the Transformer architecture. Concretely, for each head, visual representations $\tilde{\mathbf{X}}_{vis}$ are treated as queries, while the textual representations $\tilde{\mathbf{X}}_{tex}$ are treated as keys and values, which can be formulated as:
\begin{equation}
\label{visual}
    \mathbf{Z}_{tex}=\mathrm{softmax}(\frac{\mathbf{Q}_{tex}\mathbf{K}_{tex}^\top}{\sqrt{t}})\mathbf{V}_{tex}
\end{equation}
where $\mathbf{Q}_{tex}, \mathbf{K}_{tex}$, and $\mathbf{V}_{tex}$ are query vectors, key vectors, and value vectors obtained by applying the corresponding linear transformations. 
$\mathbf{Z}_{tex}\in\mathbb{R}^{k\times t}$ is the derived visual-enhanced sentence representation, where $k$ is the length of all visual blocks. After performing Eq.\ref{visual}, the sentence explicitly interacts with the associated image.

\paragraph{Sentence-rectified Visual Representation.}
Furthermore, it is important to note that each visual block may have a close association with multiple input words. To gain a deeper understanding of the relationship between images and sentences, it is necessary to align the semantic information of each visual block with its relevant words. Additionally, we aim to impose certain constraints on the image through corresponding sentences to alleviate irrelevant associations. To achieve this, we use the following Eq.\ref{language} to obtain sentence-rectified visual representations, which can complement the previously obtained visual-enhanced sentence representations. Specifically, employing the multi-head attention mechanism, the sentence representation $\tilde{\mathbf{X}}_{tex}$ acts as queries, while the visual representation $\tilde{\mathbf{X}}_{vis}$ serves as keys and values. The procedure can be formulated as follows:
\begin{equation}
\label{language}
        \mathbf{Z}_{vis}=\mathrm{softmax}(\frac{\mathbf{Q}_{vis}\mathbf{K}^\top_{vis}}{\sqrt{t}})\mathbf{V}_{vis}
\end{equation}
where $\mathbf{Q}_{vis}, \mathbf{K}_{vis}$, and $\mathbf{V}_{vis}$ are also query vectors, key vectors, and value vectors derived by performing the corresponding linear transformations. 
$\mathbf{Z}_{vis}\in\mathbb{R}^{l\times t}$ is the obtained sentence-rectified representation, which can be interpreted as constraining visual semantics from a textual perspective, where $l$ is the length of all the tokens involved in the process.

Afterwards, we merge $\mathbf{Z}_{tex}$ and $\mathbf{Z}_{vis}$ by concatenation and apply a fully connected layer to serve as the output of the image-sentence interaction module. This process can be denoted as:
\begin{equation}
    \mathbf{Z}_{int}=\phi(\mathbf{Z}_{tex}||\mathbf{Z}_{vis})
\end{equation}
where $\phi(\cdot)$ is another linear projector that projects visual-enhanced sentence representation and sentence-rectified visual representation into an interaction space, \textit{i.e.}, $\phi:\mathbb{R}^{k+l}\rightarrow\mathbb{R}^{l}$. $\mathbf{Z}_{int}\in \mathbb{R}^{l \times t}$ denotes the generated final embeddings of the full interaction of the two modalities. By incorporating information from both modalities, the resulting embeddings offer a more comprehensive representation compared to single-modal input. 
\subsection{Image-sentence Fusion}
Directly using the generated interactive embeddings may introduce potential bias and lead to suboptimal results. To alleviate this, we further design an image-sentence fusion module that applies an adaptive fusion strategy based on a gating mechanism to obtain more precise sentence-level representations and deduce the spread of errors caused by visual bias. This strategy requires simultaneously leveraging pure textual semantic representations $\tilde{\mathbf{X}}_{tex}$ and interaction representations $\mathbf{Z}_{int}$ obtained from the image-sentence interaction module. The specific architecture of the proposed module is illustrated in Fig. \ref{framework}.

We first compute the attention coefficients $\alpha$ using pure semantic representation to update the interaction representations. Then, we obtain the attention coefficients $\beta$ based on the updated interaction representations $\hat{\mathbf{Z}}_{int}$ and utilize them to obtain the updated semantic representation $\hat{\mathbf{X}}_{tex}$. This process can be formulated as follows:
\begin{equation}
    \begin{aligned}
        &\alpha=\tanh[(\tilde{\mathbf{X}}_{tex}||\mathbf{Z}_{int})\mathbf{W}_\alpha+b_\alpha] \\
        &\hat{\mathbf{Z}}_{int}=\mathbf{Z}_{int}\odot\text{softmax}(\alpha\mathbf{W}_z+b_z)  \\
        &\beta=\tanh[(\hat{\mathbf{Z}}_{int}||\tilde{\mathbf{X}}_{tex})\mathbf{W}_\beta+b_\beta] \\
        &\hat{\mathbf{X}}_{tex}=\tilde{\mathbf{X}}_{tex}\odot\mathrm{softmax}(\beta\mathbf{W}_{x}+b_x) 
    \end{aligned}
\end{equation}
where $\{\mathbf{W}_\alpha$, $\mathbf{W}_\beta\}\in \mathbb{R}^{2t\times1}$ and $\{\mathbf{W}_z$, $\mathbf{W}_x\}\in \mathbb{R}^{1\times t}$ are weight matrices. $\{b_\alpha, b_z, b_\beta, b_x\}$ are the corresponding biases. $||$ and $\odot$ denote the concatenation operation and element-wise multiplication, respectively.

Next, based on the gate mechanism, we design an approach that adaptively captures and fuses valuable and informative features from updated semantic and interaction representations, which is defined as follows:

\begin{equation}
    \begin{aligned}
    \label{fusion}g&=\sigma((\hat{\mathbf{X}}_{tex}||\hat{\mathbf{Z}}_{int})\mathbf{W}_g+b_g) \\
    \mathbf{U}&=g\cdot \hat{\mathbf{X}}_{tex}+(1-g)\cdot\hat{\mathbf{Z}}_{int}
    \end{aligned}
\end{equation}
where $g$ is the calculated gating coefficient that dynamically controls the contribution of different representations. $\mathbf{W}_g\in\mathbb{R}^{2t\times1}$ denotes the trainable parameter. $b_g$ is the bias. $\sigma(\cdot)$ represents the sigmoid function. With Eq.\ref{fusion}, we can obtain the fusion feature $\mathbf{U}\in\mathbb{R}^{l\times t}$, which adaptively integrates semantic and interaction representations.

Finally, as we mentioned before, considering the potential noise in an image, it is unreasonable to align such noise with visual blocks and sentences. To address this issue, a filtering mechanism is required to selectively leverage the fusion feature $\mathbf{U}$. If incorporating $\mathbf{U}$ enhances  the model's ability to understand a sentence, then both the fusion and original semantic features are absorbed by the filter; otherwise, it filters out fusion information. Concretely, this filtering procedure can be expressed as follows:
\begin{equation}
    \begin{aligned}
        h&=\sigma((\mathbf{U}||\tilde{\mathbf{X}}_{tex})\mathbf{W}_h+b_h) \\
        \mathbf{R}&=h \odot \tanh(\mathbf{U}\mathbf{W}_r+b_r)
    \end{aligned}
\end{equation}
where $\mathbf{W}_h\in \mathbb{R}^{2t\times 1}$ and $\mathbf{W}_r\in \mathbb{R}^{t\times t}$ are weight parameters. $b_h$ and $b_u$ are the biases. $h\in \mathbb{R}^{l}$ determines which fusion information needs to be retained. $\mathbf{R}\in \mathbb{R}^{l\times t}$ denotes the filtered representations obtained by the filtering mechanism for NLI tasks.

The classic pre-trained model based on the Transformer architecture adopts a multi-head attention mechanism. Each attention head eventually acquires an attention output $\hat{\mathbf{A}}$. 
However, in our designed scenario-guided adapter, we replace the filtered features $\mathbf{A}$ with the features $\hat{\mathbf{A}}$ obtained by the attention mechanism in pre-trained models. This ensures that the model can explicitly incorporate scenario information, thereby more comprehensively understanding sentences and alleviating word vagueness problems.
\subsection{Model Training}
After obtaining the filtered representation $\mathbf{R}$, we input it into the remaining parts of the pre-trained model based on Transformer architecture, such as layer regularization and feed-forward neural network layers, to obtain the final sentence semantic vector $\mathbf{F}\in \mathbb{R}^{\ell\times t}$ for NLI tasks, where $\ell$ represents the number of training examples.

Additionally, the objective function of this task is the cross-entropy function, denoted as:
\begin{equation}
\begin{aligned}
\hat{Y} &= \text{softmax}(\mathbf{F}\mathbf{W}_f) \\
        \mathcal{L}&=-\sum\nolimits_{i=1}^{\ell} y_i \log \hat{y}_i + (1-y_i)\log(1-\hat{y}_i)
\end{aligned}
\end{equation}
where $\hat{Y}$ is the output probability derived by a linear classifier parameterized by $\mathbf{W}_f \in \mathbb{R}^{t\times 3}$.
\section{Experiment}
\paragraph{Dataset.} We employ three datasets widely used by previous methods to validate the effectiveness of our proposed scenario-guided adapter. The statistical information of these datasets can be found in Table \ref{dataset}. The brief introduction to the datasets is provided below.
(I) \textbf{SNLI} \cite{bowman2015large} is a large-scale dataset designed for NLI tasks, where premises are summarized from photo captions in the Flickr30k and hypotheses generated by humans. \textit{Each instance in the dataset corresponds to an image, which provides scenario information}. (II) \textbf{SNLI-hard} \cite{gururangan2018annotation} is built upon the SNLI dataset, but it excludes examples from the original test set that have annotation artifacts. (III) \textbf{SNLI-lexical} \cite{glockner2018breaking} is also based on the SNLI. However, it creates a new test set that requires simple lexical inference. Note that SNLI-hard and SNLI-lexical have the same training and validation sets as SNLI.
\begin{table*}[ht]
\centering
\begin{tabular}{c|cccc|cc}
\hline
\multirow{2}{*}{Dataset} & \multirow{2}{*}{Entailment} & \multirow{2}{*}{Neutral} & \multirow{2}{*}{Contradiction} & \multirow{2}{*}{Total} & \multicolumn{2}{c}{Avg.word} \\ \cline{6-7} 
                         &                             &                          &                                &                        & premise     & hypothesis     \\ \hline
SNLI                     & 3,368                       & 3,237                    & 3,219                          & 10,000                 & 13.9        & 7.5            \\
SNLI-hard                & 1,058                       & 1,135                    & 1,068                          & 3,261                   & 13.8        & 7.7            \\
SNLI-lexical             & 982                         & 7,164                    & 47                             & 8,193                   & 11.4        & 11.6           \\ \hline
\end{tabular}%
\caption{The statistics in the test set of the evaluated datasets.}
\label{dataset}
\end{table*}

\paragraph{Baselines.} We introduce four selected categories of baseline models for comparison. 
(I) \textit{Classical sentence-based neural models} include \textbf{LSTM} \cite{hochreiter1997long}, \textbf{ESIM} \cite{chen2017enhanced}, \textbf{CAFE} \cite{tay2018compare}, and \textbf{CSRAN} \cite{tay2018co}. (II) \textit{Pre-trained language-based models} contain BERT \cite{devlin2019bert}, RoBERTa \cite{liu2019roberta}, \textbf{UERBERT} \cite{xia2021using}, \textbf{SemBERT} \cite{zhang2020semantics}, and \textbf{MT-DNN} \cite{liu2019multi}. Here, we use the base and large versions of BERT and RoBERTa models, and for the remaining models, we adopt BERT-base as the encoder. (III) \textit{Large language models} include \textbf{Bloom-7.1B} \cite{muennighoff2022crosslingual} and \textbf{Llama2-7B} \cite{touvron2023llama}. Due to our limited computational resources, we choose the approximately 7B version to the best of our ability. (IV) \textit{Multimodal models} consist of \textbf{NIC} \cite{vinyals2015show}, \textbf{m-RNN} \cite{mao2014deep}, \textbf{IEMLRN} \cite{zhang2018image}, \textbf{MIESR} \cite{zhang2019multilevel}, \textbf{VisualBERT} \cite{li2019visualbert}, and \textbf{CLIP} \cite{radford2021learning}. 

\paragraph{Implementation Details.} In the default experimental setting, we use a pre-trained ResNet-50 as the image encoder to initialize the visual representation $\mathbf{X}_{vis}$, which is later frozen during training. The obtained image embeddings $\mathbf{X}_{vis}$ and sentence token embeddings $\mathbf{X}_{tex}$ are input into the bottom Transformer block of the pre-trained BERT-base model after linear projection. The BERT-base model is fine-tuned during training. Moreover, we use the AdamW optimizer with learning rate values of \{1e-5, 2e-5, 3e-5, 5e-5\}. The warm-up and weight decay are set as 0.1 and 1e-8, respectively. The batch size is determined by grid search in \{16, 32, 64\}. Additionally, the dropout is within the range of \{0.1, 0.2, 0.3\}. Meanwhile, we apply gradient clipping within \{7.0, 10.0, 15.0\} to prevent gradient explosion. We adopt three widely used evaluation metrics: accuracy (\textit{abbr}. Acc), micro-precision (\textit{abbr}. P), and micro-recall (\textit{abbr}. R). 
\section{Result}
\begin{table*}[ht]
\centering
\tiny
\resizebox{0.9\textwidth}{!}{%
\begin{tabular}{c|ccc|ccc|ccc}
\hline
\multirow{2}{*}{Model} &
  \multicolumn{3}{c|}{SNLI} &
  \multicolumn{3}{c|}{SNLI-hard} &
  \multicolumn{3}{c}{SNLI-lexical} \\ \cline{2-10} 
        & Acc   & P     & R     & Acc   & P     & R     & Acc   & P     & R     \\ \hline
LSTM    & 80.62 & 80.49 & 80.55 & 58.53 & 58.60 & 58.49 & 52.31 & 55.59 & 52.30 \\
ESIM    & 88.02 & 88.06 & 88.05 & 71.32 & 71.66 & 71.29 & 65.60 & 68.19 & 65.59 \\
CAFE    & 88.52 & 88.59 & 88.39 & 72.15 & 72.35 & 72.14 & 66.17 & 69.73 & 66.17 \\
CSRAN   & 88.71 & 88.81 & 88.65 & 72.51 & 72.79 & 72.49 & 66.71 & 70.07 & 66.70 \\ \hline
BERT$_\text{base}$    & 90.66 & 90.43 & 90.46 & 80.50 & 80.57 & 80.49 & 92.62 & 95.41 & 92.63 \\
RoBERTa$_\text{base}$    & 90.69 & 90.41 & 90.52 & 80.53 & 80.59 & 80.46 & 92.66 & 95.45 & 92.66 \\
BERT$_\text{large}$    & 91.06 & 90.81 & 90.83 & 80.90 & 80.62 & 80.55 & 92.92 & 95.91 & 92.71 \\
RoBERTa$_\text{large}$    & 91.23 & 90.49 & 90.52 & 80.82 & 80.66 & 80.51 & 92.65 & 95.46 & 92.72 \\
UERBERT & 90.78 & 90.61 & 90.52 & 80.62 & 80.59 & 80.60 & 92.73 & 95.61 & 92.71 \\
SemBERT & 90.90 & 90.56 & 90.59 & 81.06 & 81.15 & 81.05 & 92.81 & 95.72 & 92.80 \\
MT-DNN  & 91.06 & 90.86 & 90.78 & 81.19 & 81.22 & 81.19 & 92.95 & 95.59 & 93.05 \\ \hline
Bloom-7.1B  & 91.20 & 91.16 & 91.18 & 82.02 & 81.92 & 82.01 & 95.26 & 97.79 & 95.15 \\
Llama2-7B  & 91.58 & 91.86 & 91.78 & 82.15 & 82.12 & 82.15 & 96.05 & 97.90 & 96.00 \\ \hline
NIC     & 84.75 & 84.59 & 84.62 & 63.59 & 63.52 & 63.57 & 67.12 & 72.15 & 67.15 \\
m-RNN   & 85.16 & 84.92 & 84.95 & 64.92 & 64.77 & 64.90 & 69.41 & 74.57 & 69.42 \\
IEMLRN  & 87.52 & 87.30 & 87.35 & 75.43 & 75.36 & 75.40 & 78.15 & 83.15 & 78.16 \\
MIESR   & 87.83 & 87.52 & 87.69 & 76.81 & 76.75 & 76.72 & 78.72 & 83.62 & 78.69 \\ 
VisualBERT   & 91.66 & 91.32 & 91.29 & 82.09 & 82.25 & 81.96 & 93.02 & 95.96 & 93.22 \\ 
CLIP   & 92.02 & 91.92 & 92.06 & 82.49 & 82.95 & 82.93 & 93.15 & 96.65 & 93.62 \\ \hline
ScenaFuse$_{\text{bert-base}}$ &
92.16 &
92.06 &
92.12 &
83.25 &
83.16 &
83.20 &
94.05 &
97.07 &
94.04 \\ 
ScenaFuse$_{\text{bert-large}}$ &
  92.69 &
  92.62 &
  92.66 &
  83.96 &
  83.76 &
  83.60 &
  94.75 &
  97.65 &
  94.52 \\
ScenaFuse$_{\text{bloom-7.1B}}$ &
  93.11 &
  93.06 &
  93.16 &
  84.52 &
  84.50 &
  84.46 &
  96.62 &
  98.66 &
  96.60 \\
ScenaFuse$_{\text{llama2-7B}}$ &
  \textbf{93.19} &
  \textbf{93.32} &
  \textbf{93.26} &
  \textbf{84.86} &
  \textbf{84.82} &
  \textbf{84.66} &
  \textbf{97.25} &
  \textbf{98.75} &
  \textbf{97.12} \\
  \hline
\end{tabular}%
}
\caption{Evaluation performance (\%) of various models on three datasets.}
\label{result}
\end{table*}
\paragraph{Performance Comparison.} We 
present the experimental results in Table \ref{result}. In-depth understanding based on the quantitative results are provided as follows.

\noindent $\bullet$ We observe that ScenaFuse outperforms all other models on all datasets, providing substantial evidence of its effectiveness in handling NLI tasks. This outcome could be attributed to the designed scene-guided adapter consisting of an image-sentence interaction module and an image-sentence fusion module. The model incorporates scene information, which previous models did not fully utilize, into the analysis, thereby alleviating the inherent vagueness of sentences and enabling a more comprehensive understanding of their semantic information. Furthermore, the introduction of the image-sentence interaction module allows visual representations from images to enrich sentence semantics, while textual representations from sentences can rectify visual information, benefiting each other. Next, the image-sentence fusion module deeply integrates the information from both modalities to obtain more precise sentence representations.

\noindent $\bullet$ We observe that ScenaFuse can effectively integrate with recent popular large language models (LLMs), such as Bloom and Llama2. This integration significantly improves the model performance compared to those based on other pre-trained language models. This phenomenon is expected, as LLMs are trained on massive high-quality data and possess excellent text understanding capabilities. Furthermore, we find that our model with LLMs still exhibits certain improvements compared to using LLMs alone. Because they ignore scenario information, which can guide words to produce more informative representations. 

\noindent $\bullet$ We find that sentence-based neural and multimodal models show varying degrees of decline in performance on the SNLI-hard and SNLI-lexical datasets. This finding indicates that the previously proposed NLI models' success is predominantly limited to simple examples, and their ability to recognize text entailment may not be as proficient as anticipated. When the dataset becomes challenging, \textit{i.e.}, removing annotation artifacts in the hypothesis (SNLI-hard) and requiring identifying specific words in the sentences (SNLI-lexical), it poses difficulties for all models. 

\begin{table*}[ht]
\centering
\begin{tabular}{c|ccc|ccc|ccc}
\hline
\multirow{2}{*}{Model} & \multicolumn{3}{c|}{SNLI} & \multicolumn{3}{c|}{SNLI-hard} & \multicolumn{3}{c}{SNLI-lexical} \\ \cline{2-10} 
          & Acc   & P     & R     & Acc   & P     & R     & Acc   & P     & R     \\ \hline
$\text{BERT}_\text{base}$ & 90.66 & 90.43 & 90.46 & 80.50 & 80.57 & 80.49 & 92.62 & 95.41 & 92.63 \\
$\text{ScenaFuse}_\text{rep}$ & 80.16 & 80.32 & 80.12 & 70.31 & 69.55 & 69.92 & 79.10 & 77.16 & 77.22 \\
$\text{ScenaFuse}_\text{sub}$  & 90.72 & 90.49 & 90.52 & 80.82 & 80.66 & 80.51 & 92.65 & 95.46 & 92.72 \\
ScenaFuse  & 92.16 & 92.06 & 92.12 & 83.25 & 83.16 & 83.20 & 94.05 & 97.07 & 94.04 \\ \hline
\end{tabular}%
\caption{The results (\%) of two model variants on all datasets.}
\label{sub}
\end{table*}

\paragraph{Further Discussion.} It is possible that someone may question whether scenario could directly replace the premise. If so, we only need to leverage appropriate scenarios and hypotheses to assess inferential relations. To address this doubt, we conduct an experiment where we remove all premise data from the dataset and denote the model variant as $\text{ScenaFuse}_\text{rep}$. From Table \ref{sub}, we find that the model performance on the removed premises shows a serious degradation. Although the premises summarize the content of the scenario, there are still significant differences between them. 
Scenarios may serve as reference information to enhance semantic understanding of sentences, but cannot replace premises. Because it may contain noise irrelevant to the core semantics of the sentences. Our work primarily focuses on inference relations between sentences and differs from the work on image and sentence retrieval.

Moreover, we further explore whether the scenario be substituted with the corresponding sentences. We note that the premises are extracted from photo captions in the Flickr30k corpus, indicating that the premise is a summary of the photo. Therefore, this is equivalent to generating a synonymous sentence of the original premise. 
We use the synonym substitution to generate semantically identical sentences for each premise. Then, we combine the original and the replaced premises into an extended premise and feed it into BERT for training. This model variant is denoted as $\text{ScenaFuse}_\text{sub}$. According to Table \ref{sub}, when the corresponding sentences replace the scenario, there is some performance improvement compared to BERT, but it is far behind our model. 

\begin{table}[ht]
\centering
\tiny
\resizebox{0.46\textwidth}{!}{%
\begin{tabular}{c|ccc}
\hline
Model             & SNLI          & SNLI-hard & SNLI-lexical \\ \hline
\textit{w/o ISI}  & 90.66              &   80.50        &  92.62            \\
\textit{w/o VESR} & 90.72              &   81.22        &  92.91            \\
\textit{w/o SRVR} & 90.85              &   81.36        &  93.02            \\
\textit{w/o ISF}  & 91.19              &   81.95        & 93.25             \\
\textit{w/o GM}   & 91.39              &   82.26        & 93.52             \\
\textit{w/o FM}   & 91.62              &   82.66        & 93.69             \\
ScenaFuse         & \textbf{92.16} &   \textbf{83.25}        &  \textbf{94.05}            \\ \hline
\end{tabular}%
}
\caption{The ablation results (\%) for accuracy on all datasets.}
\label{ablation}
\end{table}


\paragraph{Ablation Study.} We conduct a series of ablation studies on the evaluation dataset by designing different model variants to investigate the impact of various model components on experimental results. (I) \textit{w/o ISI}: We remove the image-sentence interaction (ISI) module and replace it with the classic multi-head attention mechanism of Transformers, which is equivalent to a complete lack of image information. (II) \textit{w/o VESR}: We eliminate the visual-enhanced sentence representation (VESR) and only retain the sentence-rectified representation in the ISI module. (III) \textit{w/o SRVR}: We eliminate the sentence-rectified representation (SRVR) and only retain the visual-enhanced sentence representation in the ISI module. (IV) \textit{w/o ISF}: We remove the image-sentence fusion (ISF) module and replace it with a simple concatenation. (V) \textit{w/o GM}: We remove the gate mechanism (GM) of the ISF module and replace it with an average of updated semantic and interaction representations. (VI) \textit{w/o FM}: We remove the filtering mechanism (FM) of the ISF module and replace it with the average of fusion features and sentence representations. 

We reach the following conclusions based on the results presented in Table \ref{ablation}. \textit{Firstly}, all the designed components make important contributions to the model, and removing any of them results in varying degrees of performance degradation. \textit{Secondly}, removing the entire ISI leads to the most significant decline in performance, indicating that scenario information can assist in sentence semantic understanding. \textit{Thirdly}, removing either VESR or SRVR leads to a performance decline, indicating that the two modules benefit each other and only appear together to learn better word embeddings. \textit{Fourth}, when the whole ISF module is deleted, performance drops due to the gap between interaction vectors and sentence vectors. Besides, removing the GM or FM of the ISF module also results in performance degradation due to potential noise interference in the image.

\section{Conclusion}
We present ScenaFuse, which is a novel model that addresses the word ambiguity and vagueness problem in NLI tasks by incorporating scenario information explicitly through visual inputs. We utilize pre-trained language models and introduce an image-sentence interaction module to extract features from both modalities. These features are then made to interact using Transformer layers. Additionally, we incorporate an image-sentence fusion module that is specially designed to adaptively fuse the learned features from images and corresponding sentences. Extensive experiments demonstrate that our proposed ScenaFuse outperforms previous competitive methods on three datasets.

\section*{Acknowledgments}
This work is supported in part by funds from the National Key Research and Development Program of China (No. 2021YFF1201200), the National Natural Science Foundation of China (No. 62172187 and No. 62372209). Fausto Giunchiglia’s work is funded by European Union’s Horizon 2020 FET Proactive project (No. 823783).

\section*{Contribution Statement}
Y.H.L. and M.Y.L. designed and developed the
method and analysed the data. Y.H.L., M.Y.L., and
X.Y.F. drafted the paper. D.L., F.G., X.M.L., and L.H. revised the
paper. X.Y.F., and R.C.G. supervised the project and contributed
to the conception of the project. 
\bibliographystyle{named}
\bibliography{ijcai24}

\begin{thebibliography}{}

\bibitem[\protect\citeauthoryear{Antol \bgroup \em et al.\egroup }{2015}]{antol2015vqa}
Stanislaw Antol, Aishwarya Agrawal, Jiasen Lu, Margaret Mitchell, Dhruv Batra, C~Lawrence Zitnick, and Devi Parikh.
\newblock Vqa: Visual question answering.
\newblock In {\em CVPR}, 2015.

\bibitem[\protect\citeauthoryear{Bowman \bgroup \em et al.\egroup }{2015}]{bowman2015large}
Samuel Bowman, Gabor Angeli, Christopher Potts, and Christopher~D Manning.
\newblock A large annotated corpus for learning natural language inference.
\newblock In {\em EMNLP}, 2015.

\bibitem[\protect\citeauthoryear{Chen \bgroup \em et al.\egroup }{2017}]{chen2017enhanced}
Qian Chen, Xiaodan Zhu, Zhen-Hua Ling, Si~Wei, Hui Jiang, and Diana Inkpen.
\newblock Enhanced lstm for natural language inference.
\newblock In {\em ACL}, 2017.

\bibitem[\protect\citeauthoryear{Conneau \bgroup \em et al.\egroup }{2017a}]{conneau2017supervised}
Alexis Conneau, Douwe Kiela, Holger Schwenk, Lo{\"\i}c Barrault, and Antoine Bordes.
\newblock Supervised learning of universal sentence representations from natural language inference data.
\newblock In {\em EMNLP}, 2017.

\bibitem[\protect\citeauthoryear{Conneau \bgroup \em et al.\egroup }{2017b}]{conneau-etal-2017-supervised}
Alexis Conneau, Douwe Kiela, Holger Schwenk, Lo{\"\i}c Barrault, and Antoine Bordes.
\newblock Supervised learning of universal sentence representations from natural language inference data.
\newblock In {\em EMNLP}, 2017.

\bibitem[\protect\citeauthoryear{Dagan \bgroup \em et al.\egroup }{2006}]{dagan2006pascal}
Ido Dagan, Oren Glickman, and Bernardo Magnini.
\newblock The pascal recognising textual entailment challenge.
\newblock In {\em First PASCAL Machine Learning Challenges Workshop}, 2006.

\bibitem[\protect\citeauthoryear{Devlin \bgroup \em et al.\egroup }{2019}]{devlin2019bert}
Jacob Devlin, Ming-Wei Chang, Kenton Lee, and Kristina Toutanova.
\newblock Bert: Pre-training of deep bidirectional transformers for language understanding.
\newblock In {\em NAACL}, 2019.

\bibitem[\protect\citeauthoryear{Falke \bgroup \em et al.\egroup }{2019}]{falke2019ranking}
Tobias Falke, Leonardo~FR Ribeiro, Prasetya~Ajie Utama, Ido Dagan, and Iryna Gurevych.
\newblock Ranking generated summaries by correctness: An interesting but challenging application for natural language inference.
\newblock In {\em ACL}, 2019.

\bibitem[\protect\citeauthoryear{Gibson \bgroup \em et al.\egroup }{2019}]{gibson2019efficiency}
Edward Gibson, Richard Futrell, Steven~P Piantadosi, Isabelle Dautriche, Kyle Mahowald, Leon Bergen, and Roger Levy.
\newblock How efficiency shapes human language.
\newblock {\em Trends in Cognitive Sciences}, 23(5):389--407, 2019.

\bibitem[\protect\citeauthoryear{Glockner \bgroup \em et al.\egroup }{2018}]{glockner2018breaking}
Max Glockner, Vered Shwartz, and Yoav Goldberg.
\newblock Breaking nli systems with sentences that require simple lexical inferences.
\newblock In {\em ACL}, 2018.

\bibitem[\protect\citeauthoryear{Guan \bgroup \em et al.\egroup }{2021}]{liu2021vplag}
Renchu Guan, Yonghao Liu, Xiaoyue Feng, and Ximing Li.
\newblock Paper-publication prediction with graph neural networks.
\newblock In {\em CIKM}, 2021.

\bibitem[\protect\citeauthoryear{Gururangan \bgroup \em et al.\egroup }{2018}]{gururangan2018annotation}
Suchin Gururangan, Swabha Swayamdipta, Omer Levy, Roy Schwartz, Samuel Bowman, and Noah~A Smith.
\newblock Annotation artifacts in natural language inference data.
\newblock In {\em NAACL}, 2018.

\bibitem[\protect\citeauthoryear{Haim \bgroup \em et al.\egroup }{2006}]{haim2006second}
R~Bar Haim, Ido Dagan, Bill Dolan, Lisa Ferro, Danilo Giampiccolo, Bernardo Magnini, and Idan Szpektor.
\newblock The second pascal recognising textual entailment challenge.
\newblock In {\em Proceedings of the Second PASCAL Challenges Workshop on Recognising Textual Entailment}, 2006.

\bibitem[\protect\citeauthoryear{He \bgroup \em et al.\egroup }{2016}]{he2016deep}
Kaiming He, Xiangyu Zhang, Shaoqing Ren, and Jian Sun.
\newblock Deep residual learning for image recognition.
\newblock In {\em CVPR}, 2016.

\bibitem[\protect\citeauthoryear{Hochreiter and Schmidhuber}{1997}]{hochreiter1997long}
Sepp Hochreiter and J{\"u}rgen Schmidhuber.
\newblock Long short-term memory.
\newblock {\em Neural Computation}, 9(8):1735--1780, 1997.

\bibitem[\protect\citeauthoryear{Li \bgroup \em et al.\egroup }{2019}]{li2019visualbert}
Liunian~Harold Li, Mark Yatskar, Da~Yin, Cho-Jui Hsieh, and Kai-Wei Chang.
\newblock Visualbert: A simple and performant baseline for vision and language.
\newblock {\em arXiv preprint arXiv:1908.03557}, 2019.

\bibitem[\protect\citeauthoryear{Li \bgroup \em et al.\egroup }{2024}]{li2024simple}
Mengyu Li, Yonghao Liu, Fausto Giunchiglia, Xiaoyue Feng, and Renchu Guan.
\newblock Simple-sampling and hard-mixup with prototypes to rebalance contrastive learning for text classification.
\newblock {\em arxiv preprint arXiv:2405.11524}, 2024.

\bibitem[\protect\citeauthoryear{Liu \bgroup \em et al.\egroup }{2019a}]{liu2019multi}
Xiaodong Liu, Pengcheng He, Weizhu Chen, and Jianfeng Gao.
\newblock Multi-task deep neural networks for natural language understanding.
\newblock In {\em ACL}, 2019.

\bibitem[\protect\citeauthoryear{Liu \bgroup \em et al.\egroup }{2019b}]{liu2019roberta}
Yinhan Liu, Myle Ott, Naman Goyal, Jingfei Du, Mandar Joshi, Danqi Chen, Omer Levy, Mike Lewis, Luke Zettlemoyer, and Veselin Stoyanov.
\newblock Roberta: A robustly optimized bert pretraining approach.
\newblock {\em arXiv preprint arXiv:1907.11692}, 2019.

\bibitem[\protect\citeauthoryear{Liu \bgroup \em et al.\egroup }{2021}]{liu2021deep}
Yonghao Liu, Renchu Guan, Fausto Giunchiglia, Yanchun Liang, and Xiaoyue Feng.
\newblock Deep attention diffusion graph neural networks for text classification.
\newblock In {\em EMNLP}, 2021.

\bibitem[\protect\citeauthoryear{Liu \bgroup \em et al.\egroup }{2022}]{liu2022few}
Yonghao Liu, Mengyu Li, Ximing Li, Fausto Giunchiglia, Xiaoyue Feng, and Renchu Guan.
\newblock Few-shot node classification on attributed networks with graph meta-learning.
\newblock In {\em SIGIR}, 2022.

\bibitem[\protect\citeauthoryear{Liu \bgroup \em et al.\egroup }{2023a}]{liulocal}
Yonghao Liu, Mengyu~Li Di~Liang, Fausto Giunchiglia, Ximing Li, Sirui Wang, Wei Wu, Lan Huang, Xiaoyue Feng, and Renchu Guan.
\newblock Local and global: Temporal question answering via information fusion.
\newblock In {\em IJCAI}, 2023.

\bibitem[\protect\citeauthoryear{Liu \bgroup \em et al.\egroup }{2023b}]{liu2023time}
Yonghao Liu, Di~Liang, Fang Fang, Sirui Wang, Wei Wu, and Rui Jiang.
\newblock Time-aware multiway adaptive fusion network for temporal knowledge graph question answering.
\newblock In {\em ICASSP}, 2023.

\bibitem[\protect\citeauthoryear{Liu \bgroup \em et al.\egroup }{2024a}]{liu2024simple}
Yonghao Liu, Lan Huang, Bowen Cao, Ximing Li, Fausto Giunchiglia, Xiaoyue Feng, and Renchu Guan.
\newblock A simple but effective approach for unsupervised few-shot graph classification.
\newblock In {\em WWW}, 2024.

\bibitem[\protect\citeauthoryear{Liu \bgroup \em et al.\egroup }{2024b}]{liu2024improved}
Yonghao Liu, Lan Huang, Fausto Giunchiglia, Xiaoyue Feng, and Renchu Guan.
\newblock Improved graph contrastive learning for short text classification.
\newblock In {\em AAAI}, 2024.

\bibitem[\protect\citeauthoryear{MacCartney}{2009}]{maccartney2009natural}
Bill MacCartney.
\newblock {\em Natural language inference}.
\newblock Stanford University, 2009.

\bibitem[\protect\citeauthoryear{Mao \bgroup \em et al.\egroup }{2015}]{mao2014deep}
Junhua Mao, Wei Xu, Yi~Yang, Jiang Wang, Zhiheng Huang, and Alan Yuille.
\newblock Deep captioning with multimodal recurrent neural networks (m-rnn).
\newblock In {\em ICLR}, 2015.

\bibitem[\protect\citeauthoryear{Moon \bgroup \em et al.\egroup }{2018}]{moon2018multimodal}
Seungwhan Moon, Leonardo Neves, and Vitor Carvalho.
\newblock Multimodal named entity recognition for short social media posts.
\newblock In {\em NAACL}, 2018.

\bibitem[\protect\citeauthoryear{Muennighoff \bgroup \em et al.\egroup }{2022}]{muennighoff2022crosslingual}
Niklas Muennighoff, Thomas Wang, Lintang Sutawika, Adam Roberts, Stella Biderman, Teven~Le Scao, M~Saiful Bari, Sheng Shen, Zheng-Xin Yong, Hailey Schoelkopf, et~al.
\newblock Crosslingual generalization through multitask finetuning.
\newblock {\em arXiv preprint arXiv:2211.01786}, 2022.

\bibitem[\protect\citeauthoryear{Murty \bgroup \em et al.\egroup }{2021}]{murty2021dreca}
Shikhar Murty, Tatsunori~B Hashimoto, and Christopher~D Manning.
\newblock Dreca: A general task augmentation strategy for few-shot natural language inference.
\newblock In {\em NAACL}, 2021.

\bibitem[\protect\citeauthoryear{Nishihara \bgroup \em et al.\egroup }{2020}]{nishihara2020supervised}
Tetsuro Nishihara, Akihiro Tamura, Takashi Ninomiya, Yutaro Omote, and Hideki Nakayama.
\newblock Supervised visual attention for multimodal neural machine translation.
\newblock In {\em COLING}, 2020.

\bibitem[\protect\citeauthoryear{Pan \bgroup \em et al.\egroup }{2016}]{pan2016jointly}
Yingwei Pan, Tao Mei, Ting Yao, Houqiang Li, and Yong Rui.
\newblock Jointly modeling embedding and translation to bridge video and language.
\newblock In {\em CVPR}, 2016.

\bibitem[\protect\citeauthoryear{Pinker}{2003}]{pinker2003mind}
Steven Pinker.
\newblock {\em How the mind works}.
\newblock Penguin UK, 2003.

\bibitem[\protect\citeauthoryear{Radford \bgroup \em et al.\egroup }{2021}]{radford2021learning}
Alec Radford, Jong~Wook Kim, Chris Hallacy, Aditya Ramesh, Gabriel Goh, Sandhini Agarwal, Girish Sastry, Amanda Askell, Pamela Mishkin, Jack Clark, et~al.
\newblock Learning transferable visual models from natural language supervision.
\newblock In {\em ICML}, 2021.

\bibitem[\protect\citeauthoryear{Ramachandram and Taylor}{2017}]{ramachandram2017deep}
Dhanesh Ramachandram and Graham~W Taylor.
\newblock Deep multimodal learning: A survey on recent advances and trends.
\newblock {\em IEEE SPM}, 34(6):96--108, 2017.

\bibitem[\protect\citeauthoryear{Rockt{\"a}schel \bgroup \em et al.\egroup }{2016}]{rocktaschel2015reasoning}
Tim Rockt{\"a}schel, Edward Grefenstette, Karl~Moritz Hermann, Tom{\'a}{\v{s}} Ko{\v{c}}isk{\`y}, and Phil Blunsom.
\newblock Reasoning about entailment with neural attention.
\newblock In {\em ICLR}, 2016.

\bibitem[\protect\citeauthoryear{Simonyan and Zisserman}{2014}]{simonyan2014very}
Karen Simonyan and Andrew Zisserman.
\newblock Very deep convolutional networks for large-scale image recognition.
\newblock {\em arXiv preprint arXiv:1409.1556}, 2014.

\bibitem[\protect\citeauthoryear{Storks \bgroup \em et al.\egroup }{2019}]{storks2019recent}
Shane Storks, Qiaozi Gao, and Joyce~Y Chai.
\newblock Recent advances in natural language inference: A survey of benchmarks, resources, and approaches.
\newblock {\em arXiv preprint arXiv:1904.01172}, 2019.

\bibitem[\protect\citeauthoryear{Tay \bgroup \em et al.\egroup }{2018a}]{tay2018co}
Yi~Tay, Anh~Tuan Luu, and Siu~Cheung Hui.
\newblock Co-stack residual affinity networks with multi-level attention refinement for matching text sequences.
\newblock In {\em EMNLP}, 2018.

\bibitem[\protect\citeauthoryear{Tay \bgroup \em et al.\egroup }{2018b}]{tay2018compare}
Yi~Tay, Anh~Tuan Luu, and Siu~Cheung Hui.
\newblock Compare, compress and propagate: Enhancing neural architectures with alignment factorization for natural language inference.
\newblock In {\em EMNLP}, 2018.

\bibitem[\protect\citeauthoryear{Tong \bgroup \em et al.\egroup }{2020}]{tong2020image}
Meihan Tong, Shuai Wang, Yixin Cao, Bin Xu, Juanzi Li, Lei Hou, and Tat-Seng Chua.
\newblock Image enhanced event detection in news articles.
\newblock In {\em AAAI}, 2020.

\bibitem[\protect\citeauthoryear{Touvron \bgroup \em et al.\egroup }{2023}]{touvron2023llama}
Hugo Touvron, Louis Martin, Kevin Stone, Peter Albert, Amjad Almahairi, Yasmine Babaei, Nikolay Bashlykov, Soumya Batra, Prajjwal Bhargava, Shruti Bhosale, et~al.
\newblock Llama 2: Open foundation and fine-tuned chat models.
\newblock {\em arXiv preprint arXiv:2307.09288}, 2023.

\bibitem[\protect\citeauthoryear{Vendrov \bgroup \em et al.\egroup }{2016}]{vendrov2015order}
Ivan Vendrov, Ryan Kiros, Sanja Fidler, and Raquel Urtasun.
\newblock Order-embeddings of images and language.
\newblock In {\em ICLR}, 2016.

\bibitem[\protect\citeauthoryear{Vinyals \bgroup \em et al.\egroup }{2015}]{vinyals2015show}
Oriol Vinyals, Alexander Toshev, Samy Bengio, and Dumitru Erhan.
\newblock Show and tell: A neural image caption generator.
\newblock In {\em CVPR}, 2015.

\bibitem[\protect\citeauthoryear{Wang and Jiang}{2016}]{wang2016learning}
Shuohang Wang and Jing Jiang.
\newblock Learning natural language inference with lstm.
\newblock In {\em NAACL}, 2016.

\bibitem[\protect\citeauthoryear{Williams \bgroup \em et al.\egroup }{2018}]{williams2018broad}
Adina Williams, Nikita Nangia, and Samuel Bowman.
\newblock A broad-coverage challenge corpus for sentence understanding through inference.
\newblock In {\em NAACL}, 2018.

\bibitem[\protect\citeauthoryear{Xia \bgroup \em et al.\egroup }{2021}]{xia2021using}
Tingyu Xia, Yue Wang, Yuan Tian, and Yi~Chang.
\newblock Using prior knowledge to guide bert’s attention in semantic textual matching tasks.
\newblock In {\em WWW}, 2021.

\bibitem[\protect\citeauthoryear{Zhang \bgroup \em et al.\egroup }{2010}]{zhang2010understanding}
Yin Zhang, Rong Jin, and Zhi-Hua Zhou.
\newblock Understanding bag-of-words model: a statistical framework.
\newblock {\em IJMLC}, 1:43--52, 2010.

\bibitem[\protect\citeauthoryear{Zhang \bgroup \em et al.\egroup }{2018}]{zhang2018image}
Kun Zhang, Guangyi Lv, Le~Wu, Enhong Chen, Qi~Liu, Han Wu, and Fangzhao Wu.
\newblock Image-enhanced multi-level sentence representation net for natural language inference.
\newblock In {\em ICDM}, 2018.

\bibitem[\protect\citeauthoryear{Zhang \bgroup \em et al.\egroup }{2019}]{zhang2019multilevel}
Kun Zhang, Guangyi Lv, Le~Wu, Enhong Chen, Qi~Liu, Han Wu, Xing Xie, and Fangzhao Wu.
\newblock Multilevel image-enhanced sentence representation net for natural language inference.
\newblock {\em IEEE TSMC}, 51(6):3781--3795, 2019.

\bibitem[\protect\citeauthoryear{Zhang \bgroup \em et al.\egroup }{2020}]{zhang2020semantics}
Zhuosheng Zhang, Yuwei Wu, Hai Zhao, Zuchao Li, Shuailiang Zhang, Xi~Zhou, and Xiang Zhou.
\newblock Semantics-aware bert for language understanding.
\newblock In {\em AAAI}, 2020.

\end{thebibliography}

\end{document}